\documentclass[shortpaper]{clv3}

\usepackage{enumitem}
\usepackage{url}
\usepackage{booktabs}
\usepackage{amsmath}
\usepackage{amssymb} 
\usepackage{subcaption}
\usepackage{caption}

\DeclareMathOperator*{\argmin}{arg\,min}

\usepackage{hyperref}
\usepackage[capitalise]{cleveref}

\historydates{Submission received: 7 June 2021; revised version received: 9 December 2021; accepted for publication: 21 December 2021}

\begin{document}

\title{Annotation Curricula to Implicitly Train Non-Expert Annotators}
\runningtitle{Annotation Curricula}
\runningauthor{Lee, Klie, Gurevych}

\author{Ji-Ung Lee\thanks{Equal contribution}}
\affil{UKP Lab / TU Darmstadt}

\author{Jan-Christoph Klie\footnotemark[1]}
\affil{UKP Lab / TU Darmstadt}

\author{Iryna Gurevych}
\affil{UKP Lab / TU Darmstadt}

\maketitle

\begin{abstract}
Annotation studies often require annotators to familiarize themselves with the task, its annotation scheme, and the data domain.
This can be overwhelming in the beginning, mentally taxing, and induce errors into the resulting annotations; especially in citizen science or crowdsourcing scenarios where domain expertise is not required.
To alleviate these issues, this work proposes annotation curricula, a novel approach to implicitly train annotators. 
The goal is to gradually introduce annotators into the task by ordering instances to be annotated according to a learning curriculum.
To do so, this work formalizes annotation curricula for sentence- and paragraph-level annotation tasks, defines an ordering strategy, and identifies well-performing heuristics and interactively trained models on three existing English datasets. 
Finally, we provide a proof of concept for annotation curricula in a carefully designed user study with 40 voluntary participants who are asked to identify the most fitting misconception for English tweets about the Covid-19 pandemic.
The results indicate that using a simple heuristic to order instances can already significantly reduce the total annotation time while preserving a high annotation quality. 
Annotation curricula thus can be a promising research direction to improve data collection. 
To facilitate future research --- for instance, to adapt annotation curricula to specific tasks and expert annotation scenarios --- all code and data from the user study consisting of 2,400 annotations is made available.\footnote{\url{https://github.com/UKPLab/annotation-curriculum}.} 
\end{abstract}

\section{Introduction}
\label{introduction}
Supervised learning and, consequently, annotated corpora are crucial for many down-stream tasks to train and develop well-performing models.
Despite improvements of models trained in a semi- or unsupervised fashion~\citep{peters-etal-2018-deep,devlin-etal-2019-bert}, they still substantially benefit from labeled data~\citep{peters-etal-2019-tune,gururangan-etal-2020-dont}. 
However, labels are costly to obtain and require domain experts or a large crowd of non-expert annotators~\citep{snow-etal-2008-cheap}.

Past research has mainly investigated two approaches to reduce annotation cost and effort (often approximated by annotation time); namely, \textbf{active learning} and \textbf{label suggestions}. 
Active learning assumes that resources for annotating data are limited and aims to reduce the number of labeled instances by only annotating those which contribute most to model training~\citep{lewis-etal-1994-sequential, settles-2012-active}.
This often results in sampled instances that are more difficult to annotate, putting an increased cognitive load on annotators, and potentially leading to a lower agreement or an increased annotation time~\citep{settles-etal-2008-al}.
Label suggestions directly target annotators by providing them with suggestions from a pre-trained model. 
Although they are capable of effectively reducing the annotation time~\citep{schulz-etal-2019-analysis,klie-etal-2020-zero,beck-etal-2021-investigating}, they bear the risk of biasing annotators towards the (possibly erroneous) suggested label~\citep{fort-sagot-2010-influence}. 
Both these shortcomings render existing approaches better suited for domain-expert annotators who are less burdened by difficult annotation instances and are less prone to receiving erroneous label suggestions than non-expert annotators.
Overall, we can identify a lack of approaches that (1) are less distracting or biased than label suggestions and (2) can also ease the annotation process for non-expert annotators.
Especially the increasing popularity of large-scale, crowdsourced datasets~\citep{bowman-etal-2015-large,sakaguchi2021winogrande} further amplifies the need for training methods that can also be applied in non-expert annotator scenarios~\citep{geva-etal-2019-modeling,nie-etal-2020-adversarial,rogers-2021}.

One key element that has so far not been investigated in annotation studies is the use of a curriculum to \textit{implicitly} teach the task to annotators during annotation.
The \textbf{learning curriculum} is a fundamental concept in educational research that proposes to order exercises to match a learner's proficiency~\citep{Vygotsky78,krashen1982principles} and has even motivated training strategies for machine learning models~\citep{bengio-etal-2009-curriculum}.
Moreover, \citet{kelly2009curriculum} showed that such learning curricula can also be used to teach learners implicitly.
Similarly, the goal of \textbf{annotation curricula} (AC) is to provide an ordering of instances during annotation that is optimized for learning the task. 
We conjecture that a good annotation curriculum can implicitly teach the task to annotators --- for instance, by showing easier annotation instances before more difficult ones --- consequently reducing the cognitive strain and improving annotation speed and quality.
In contrast to active learning that may result in only sampling instances that are difficult to annotate, they explicitly emphasize the needs of a human annotator and gradually familiarize them with the annotation task.
Compared to label suggestions, they are less distracting as they do not bear the risk of providing erroneous suggestions from imperfect models which makes them well-suited for non-expert annotation scenarios.
Furthermore, AC do not require study conductors to adapt existing annotator training processes or annotation guidelines and hence, can complement their annotation project. 
To provide a first assessment for the viability of such annotation curricula, we investigate the following three research questions:

\begin{itemize}[topsep=5pt,itemsep=3pt]
	\item[\textbf{RQ1.}] Does the order in which instances are annotated impact the annotations in terms of annotation time and quality?
	\item[\textbf{RQ2.}] Do traditional heuristics and recent methods for assessing the reading difficulty already suffice to generate curricula that improve annotation time or quality?
	\item[\textbf{RQ3.}] Can the generation of annotation curricula be further alleviated by interactively trained models?
\end{itemize}

We first identify and formalize two essential parts to deploy AC: (1) a "strategy" that defines how instances should be ordered (e.g., by annotation difficulty) and (2) an "estimator" that ranks them accordingly. 
We instantiate AC with an "easy-instances-first" strategy and evaluate heuristic and interactively trained estimators on three English datasets that provide annotation time which we use as an approximation of the annotation difficulty for evaluation.
Finally, we apply our strategy and its best estimators in a carefully designed user study with 40 participants for annotating English tweets about the Covid-19 pandemic.
The study results show that the ordering in which instances are annotated can have a statistically significant impact on the outcome.
We furthermore find that annotators who receive the same instances in an optimized order require significantly less annotation time while retaining a high annotation quality.
Our contributions are:

\begin{itemize}[topsep=5pt,itemsep=3pt]
	\item[\textbf{C1.}] A novel approach for training non-expert annotators that is easy to implement and is complementary to existing annotator training approaches. 
	\item[\textbf{C2.}] A formalization of AC for sentence- and paragraph-labeling tasks with a strategy that orders instances from easy to difficult, and an evaluation for three heuristics and three interactively trained estimators.
	\item[\textbf{C3.}] A first evaluation of AC in a carefully designed user study that controls for external influences including:
	\begin{itemize}
		\item[a)] An implementation of our evaluated AC strategies and 2,400 annotations collected during our human evaluation study. 
		\item[b)] A production-ready implementation of interactive AC in the annotation framework INCEpTION ~\citep{klie-2018-inception} that can be readily deployed.
	\end{itemize}
\end{itemize}

Our evaluation of different heuristics and interactively trained models further reveals additional factors --- such as the data domain and the annotation task --- that can influence their aptitude for AC.
We thus appeal to study conductors to publish the annotation order and annotation times along with their data to allow future works to better investigate and develop task- and domain-specific AC.

\section{Related Work}
\label{rel_work}
Most existing approaches that help with data collection focus on either active learning or label suggestions.
Other researchers also investigate tackling annotation task within the context of gamification and introduce different levels of difficulty.

\paragraph{Active learning}
Active learning has widely been researched in terms of model-oriented approaches~\citep{lewis-etal-1994-sequential,roy-etal-2001-toward,gal-etal-2017-deep,siddhant-lipton-2018-deep,kirsch-etal-2019-batchbald}, data-oriented approaches~\citep{nguyen-etal-2004-precluster,zhu-etal-2008-active,huang-etal-2010-density,wang-etal-2017-active} or combinations of both~\citep{Ash2020Deep,yuan-etal-2020-cold}.
Although several works investigate annotator proficiency --- which is especially important for crowdsourcing --- their main concern is to identify noisy labels or erroneous annotators \citep{laws-etal-2011-active,fang-etal-2012-selftaught,zhang-etal-2015-active} or distribute tasks between workers of different proficiency~\citep{fang-etal-2014-alcrowd,yang_predicting_2019}.
Despite the large amount of research in active learning, only few works have considered annotation time as an additional cost variable in active learning~\citep{settles-etal-2008-al} and even found that active learning can negatively impact annotation time~\citep{martinez-alonso-etal-2015-active}.
Other practical difficulties for deploying active learning in real annotation studies stem from additional hyper-parameters that are introduced, but seldom investigated~\citep{lowell-etal-2019-practical}.
In contrast, AC also works well with simple heuristics, allowing researchers to pre-compute the order of annotated instances.

\paragraph{Label suggestions} 
Label suggestions have been considered for various annotation tasks in NLP, such as in part-of-speech tagging for low-resource languages~\citep{yimam-etal-2014-automatic}, interactive entity-linking~\citep{klie-etal-2020-zero} or identifying evidence in diagnostic reasoning~\citep{schulz-etal-2019-analysis}.
Especially for tasks that require domain-specific knowledge such as in the medical domain, label suggestions can substantially reduce the burden on the annotator~\citep{lingren2014evaluating}.
However, they also inherently pose the risk of amplifying annotation biases due to the anchoring effect~\citep{turner2016anchor}.
Whereas domain experts may be able to reliably identify wrong suggestions and provide appropriate corrections~\citep{fort-sagot-2010-influence}, this cannot be assumed for non-experts. 
This renders label suggestions a less viable solution to ease annotations in non-expert studies where incorrect label suggestions may even distract annotators from the task.
In contrast, changing the ordering in which instances are annotated by using an AC is not distracting at all.  

\paragraph{Annotation difficulty}
Although difficulty estimation is crucial in human language learning, for instance, in essay scoring~\citep{mayfield-black-2020-fine} or text completion exercises~\citep{beinborn-2014,loukina-etal-2016-textual, lee-etal-2019-manipulating}, it is difficult to achieve in annotation scenarios due to the lack of ground truth, commonly resulting in a post-annotation analysis for model training~\citep{beigman-klebanov-beigman-2014-difficult,paun-etal-2018-comparing}.
To consider the difficulty of annotated instances, a concept that has recently been explored for (annotation) games with a purpose, is \textbf{progression}.
It allows annotators to progress through the annotation study similar to a game --- by acquiring specific skills that are required to progress to the next level~\citep{sweetser2005gameflow}.
Although several works have shown the efficiency of progression in games with a purpose~\citep{Madge2019ProgressionIA,kicikoglu-etal-2020-aggregation} and even in crowdsourcing~\citep{tauchmann-2020}, this does not necessarily benefit individual workers as less skilled workers are either filtered out or asked to "train" on additional instances.
Moreover, implementing progression poses a substantial burden on researchers due to the inclusion of game-like elements (e.g., skills and levels), or at minimum, the separation of the data according to difficulty and furthermore, a repeated evaluation and reassignment of workers.
In contrast, reordering instances of a single set according to a given curriculum can already be achieved with low effort and can even be implemented complementary to progression.

\section{Annotation Curriculum}
\label{overview}
We first specify the type of annotation tasks investigated in this work, and then formalize AC with the essential components that are required for generating appropriate annotation curricula.
Finally, we instantiate an easy-instances-first strategy and define the estimators that we use to generate a respective curriculum.

\subsection{Annotation Task}
\label{annotation_task}
In this work, we focus on sentence- and paragraph-level annotation tasks that do not require any deep domain-expertise and hence, are often conducted with non-expert annotators.\footnote{We discuss AC strategies that may be better suited for domain experts in \cref{sec:limitations}.}
Such annotation tasks often use a simple annotation scheme which is limited to a small set of labels, and have been used to create datasets across various research areas, for instance, in sentiment analysis~\citep{pak-paroubek-2010-twitter-corpus}, natural language inference~\citep{bowman-etal-2015-large}, and argument mining~\citep{stab-etal-2018-argumentext}.

\paragraph{Task Formalization} 
We define an annotation task as being composed of a set of unlabeled instances $x \in \mathcal{U}$ that are to be annotated with their respective labels $y \in \mathcal{Y}$.
We focus on instances $x$ that are either a sentence or a paragraph and fully annotated by an annotator $a$. 
Note that for sequence labeling tasks such as named entity recognition, $y$ is not a single label but a vector comprised of the respective token-level labels.
However, in such tasks, annotations are still often collected for a complete sentence or paragraph at once to provide annotators with the necessary context~\citep{tomanek_timed_2009}.

\subsection{Approach}
\label{framework}

\begin{figure}
	\centering
	\includegraphics[width=0.85\linewidth ]{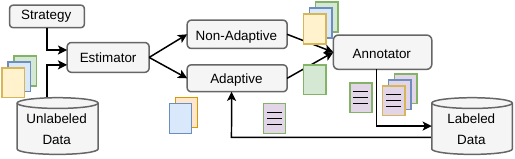}
	\caption{Annotation curricula. First, we define a strategy for ordering instances by annotation difficulty (i.e., easy-first). We then implement estimators that perform the ordering. Estimators can either be non-adaptive (e.g., heuristics) or adaptive (trained models). Finally, annotators receive instances according to the resulting curriculum.}
	\label{fig:ac}
\end{figure}

\cref{fig:ac} provides a general overview of AC.
Given a set of unlabeled instances $x \in \mathcal{U}$, we define a strategy $\mathcal{S}$ that determines the ordering in which annotated instances should be presented (easy-instances-first). 
We then specify "adaptive" and "non-adaptive" estimators $f(\cdot)$ that approximate the true annotation difficulty. 
In this work, we focus on task-agnostic estimators that can easily be applied across a wide range of tasks and leave the investigation on task-specific estimators --- that may have a higher performance but also require more implementation effort from study conductors --- for future work.\footnote{We discuss some ideas for task-specific estimators in \cref{sec:limitations}.} 
Depending on the estimator, we then order the annotated instances either beforehand (non-adaptive), or select them iteratively at each step based on the predictions of an interactively trained model (adaptive).

\paragraph{Formalization}
Ideally, an annotation curriculum that optimally introduces annotators to the task would minimize (1) annotation effort and (2) error rate (i.e., maximize annotation quality).
As the annotation error can only be obtained post-study, we can only use annotation effort, approximated by annotation time, for our formalization; however, we conjecture that minimizing annotation time may also have a positive impact on annotation quality (given that the annotators remain motivated throughout their work).
To further reduce noise factors during evaluation, we focus on annotation studies that involve a limited number of instances (in contrast to active learning scenarios that assume an abundance of unlabeled data). 
We thus formalize AC as the task of finding the optimal curriculum $\mathcal{C}^*$ out of all possible curricula $\mathcal{C}$ (i.e., permutations of $\mathcal{U}$) for a finite set of unlabeled instances $\mathcal{U}$ that minimizes the total annotation time $\mathcal{T}$; namely, the sum of individual annotation times $t_i \in \mathbb{R}^+$ for all instances $x_i\in\mathcal{U}$ with $i$ denoting the $i$-th annotated instance:
\begin{equation}\label{eq:1}
\mathcal{C}^* = \argmin_{\mathcal{C}} \sum_{i=1}^{|\mathcal{U}|} a_i(x_i | x_0 \ldots  x_{i-1})
\end{equation}
where $a_i : \mathcal{U} \rightarrow  \mathcal{T}$ describes the annotator after annotating $i-1$ instances.

\paragraph{Strategy}
Due to the large number of $n!$ possible curricula $\mathcal{C}$ resulting from $n=|\mathcal{U}|$ instances, solving \cref{eq:1} is intractable for large n  even if $a(\cdot)$ was known.
We can furthermore only assess the true effectiveness of a curriculum $\mathcal{C}$ post-study, making it impossible to find the optimal curriculum $\mathcal{C}^*$ beforehand.
We hence require a strategy $\mathcal{S} \sim \mathcal{C}^*$ that specifies how instances of  $\mathcal{U}$ should be ordered optimally. 
Similar to educational approaches, we rely on estimating the "difficulty" of an instance to generate our curriculum~\citep{taylor1953cloze,beinborn-2014,lee-etal-2019-manipulating}. 
In this work, we investigate an easy-instances-first strategy which has been shown to be a reasonable strategy in previous work~\citep{tauchmann-2020}; thereby sorting instances in ascending order according to their difficulty.
Our $\mathcal{C}^*$ is thus approximated by the ordered set $\mathcal{S} = \{x_1, \ldots, x_n | \forall x_{1 \leq i \leq n} \in \mathcal{S} : f(x_i) \leq f(x_{i+1})\}$ with $f(\cdot)$ being the difficulty estimator.

\paragraph{Non-adaptive estimators}
We define non-adaptive estimators as heuristics or pre-trained models that are not updated interactively.
The respective annotation curriculum can thus be pre-computed and does not impose any additional changes to the underlying annotation platform. 
To estimate the annotation difficulty, non-adaptive estimators define a scoring function $f_{\bar{a}} : \mathcal{U} \to \mathbb{R}$.
In this work, we evaluate non-adaptive estimators that are commonly used in readability assessment to score the reading difficulty of a text~\citep{xia-etal-2016-text,deutsch-etal-2020-linguistic}.
Although they are not capable of capturing any task-specific difficulties, they have the advantage of being applicable to a wide range of tasks with low effort for study conductors. 
The following heuristics and pre-trained models are investigated to obtain difficulty estimations for the easy-instances-first curriculum: 

\begin{description}[noitemsep,topsep=3pt,itemsep=3pt,itemindent=-1em]
	\item[Sentence Length ($\mathrm{sen}$)] The average number of tokens in the annotated instance.
	\item[Flesch-Kincaid ($\mathrm{FK}$)] A readability score based on the number of words, syllables, and sentences~\citep{kincaid1975derivation}.
	\item[Masked Language Modeling Loss ($\mathrm{mlm}$)] As shown in recent works, the losses of a masked language model may be used to obtain an assessment of text complexity~\citep{felice-buttery-2019-entropy}. 
	We use the implementation of \citet{salazar-etal-2020-masked}.
\end{description}

\paragraph{Adaptive estimators}
While simple heuristics or annotator-unaware models allow us to pre-compute annotation curricula, they do not consider any user-specific aspect that may influence the difficulty estimation~\citep{lee-etal-2020-empowering}.
Consequently, the resulting curriculum may not provide the optimal ordering for a specific annotator.
To select the instance with the most appropriate difficulty for an annotator $a_i(\cdot)$ at the $i$-th iteration, we use a model ${\theta_i(\cdot)\sim a_i(\cdot)}$ that is updated with an increasing number of annotated instances.
We conjecture that using $\theta(\cdot)$ to predict the relative difficulty --- in contrast to non-adaptive estimators that provide an absolute difficulty estimation --- may be more robust to task-specific influences as they are inherited in all instances annotated by $a(\cdot)$.
When training adaptive estimators, we use annotation time to approximate the difficulty of a specific instance due to its availability in any annotation scenario.
At iteration $i$, we thus train the model $\theta_{i}: \mathcal{L} \to \mathcal{T} \subseteq \mathbb{R}^+$ to predict the annotation times $t \in \mathcal{T}$ for all labeled instances $\hat{x} \in \mathcal{L}$. 
Similar to active learning, we now encounter a decreasing number of unlabeled instances and an increasing number of labeled instances. 
The resulting model is then used to estimate the annotation time for all unlabeled instances $x \in \mathcal{U}$. 
The resulting scoring function is now defined as $f_a : \theta_i, \mathcal{U} \to \mathbb{R}^+$. 
Finally, we select instance $x^* \in \mathcal{U}$ with the minimal rank according to $f_a$. 
\begin{equation}
x^* = \argmin_{f_a} \theta_{i}(x)
\end{equation}

\noindent
Following our strategy $\mathcal{S}$, this results in selecting instances for annotation that have the lowest predicted annotation time.
We specifically focus on regression models that can be trained efficiently in-between annotation and work robustly in low-data scenarios. 
We choose Ridge Regression ($\mathrm{RR}$), Gaussian Process Regression ($\mathrm{GP}$) and GBM Regression ($\mathrm{GBM}$).

\section{Evaluation with Existing Datasets}
\label{intrinsic}
To identify well-performing non-adaptive and adaptive estimators, we first evaluate AC on existing datasets in an offline setting.
We focus on datasets that provide annotation time which is used to approximate the annotation difficulty during evaluation (to address the lack of gold labels in actual annotation scenarios).
Following~\citet{settles-etal-2008-al}, we conjecture that instances with a higher difficulty require more time to annotate.
For comparison, we then compute the correlations between different orderings generated according to our easy-instances-first strategy using text difficulty heuristics (non-adaptive) and interactively trained models (adaptive) with the annotation time (approximated annotation difficulty).
We evaluate our estimators in two setups:

\begin{description}[noitemsep,topsep=3pt,itemsep=3pt,itemindent=-1em]
	\item[Full] We evaluate how well adaptive and non-adaptive estimators trained on the whole training set correlate with the annotation time of the respective test set (upper bound).
	\item[Adaptive] We evaluate the performance of adaptive estimators in an interactive learning scenario with simulated annotators and an increasing number of training instances.
\end{description}

\subsection{Datasets}
Overall, we identify three NLP datasets that provide accurate annotation time for individual instances along with their labels:

\begin{description}[noitemsep,topsep=3pt,itemsep=3pt,itemindent=-1em]
	\item[Muc7\textsubscript{T}] \citet{tomanek_timed_2009} extended the \textsc{Muc7} corpus that consists of annotated named entities in English Newswire articles. 
	They reannotated the data with two annotators A and B while measuring their annotation time per sentence.
	\item[SigIE]  is a collection of email signatures that was tagged by~\citet{settles-etal-2008-al} with twelve named entity types typical for email signatures such as phone number, name, and job title.
	\item[SPEC] The same authors~\citep{settles-etal-2008-al} further annotated sentences from 100 English PubMed abstracts according to their used language (speculative or definite) with three annotators.
\end{description}

\begin{table}[ht]
	\centering
	\begin{tabular}{@{}lcrrrrrrrr@{}}
		\toprule
		Name   &  Task   &  $|\mathbf{\mathcal{D}}|$ & $|\mathbf{\mathcal{D}_{train}}|$ & $|\mathbf{\mathcal{D}_{dev}}|$ & $|\mathbf{\mathcal{D}_{test}}|$  & $\mu_{|\mathbf{\mathcal{D}}|}$ & $\sigma_{|\mathbf{\mathcal{D}}|}$ &  $\mu_t$  &  $\sigma_t$  \\
		\midrule
		Muc7\textsubscript{T} A  & ST     & 3113  &  2179   &  467  &  467   &   133.7   &   70.8   &  5.4   &  3.9  \\
		Muc7\textsubscript{T} B & ST     & 3113  &  2179   &  467  &  467   &   133.7   &   70.8   &  5.2   &  4.2  \\
		SigIE  & ST     &  251  &   200   &    -   &   51   &   226.4   &  114.8   &  27.0  & 14.7  \\
		SPEC   & Cl     &  850  &   680   &   -    &  170   &   160.4   &   64.2   &  22.7  & 12.4  \\
		\bottomrule
	\end{tabular}
	\caption{Annotation task (ST for sequence tagging, Cl for classification) and the number of instances per dataset and split. $\mu_{|\mathbf{\mathcal{D}}|}$ denotes the average instance length in characters and $\mu_t$ the average annotation time. $\sigma_{|\mathbf{\mathcal{D}}|}$ and $\sigma_t$ denotes the standard deviation, respectively. Across all datasets, annotation time is reported for annotating the whole instance (i.e., not for individual entities).}
	\label{tab:datasets-overview}
\end{table}

\cref{tab:datasets-overview} provides an overview of the used datasets.
It can be seen that Muc7\textsubscript{T} is the largest corpus ($|\mathcal{D}|$), however, it is also the one that consists of the shortest instances on average ($\mu_{|\mathbf{\mathcal{D}}|}$).
Furthermore, Muc7\textsubscript{T} also has the lowest annotation times ($\mu_t$) and a low standard deviation ($\sigma_t$). 
Comparing the number of entities per instance between Muc7\textsubscript{T} (news articles) and SigIE (email signatures) shows their differences with respect to their domains with an average number of 1.3 entities ($\sigma=1.4$) in Muc7\textsubscript{T} and 5.3 entities ($\sigma=3.0$) in SigIE.
Moreover, we find that the SigIE corpus has a higher ratio of entity tokens (40.5\%) than Muc7\textsubscript{T} (8.4\%) which may explain the long annotation time.
Interestingly, the binary sentence classification task SPEC ("speculative" or "definite") also displays a substantially longer annotation time compared to Muc7\textsubscript{T} (on average, more than four times) which may also indicate a higher task difficulty or less proficiency of the involved annotators.

\paragraph{Data splits}
For Muc7\textsubscript{T}, we focus on the annotations of the first annotator Muc7\textsubscript{T} A; using Muc7\textsubscript{T} B yields similar results.
For SPEC, we use \textsc{all.dat} for our experiments.
None of the aforementioned datasets provide default splits. 
We hence create 80-20 train-test splits of SPEC and SigIE for our experiments. 
To identify the best hyper-parameters of our adaptive estimators, we split the largest corpus (Muc7\textsubscript{T}) into 70-15-15 train-dev-test splits.
All splits are published along with the code and data.

\subsection{Experimental Setup}
Our goal is to evaluate how well the ordering generated by an estimator correlates with the annotation time provided in the respective datasets.

\paragraph{Evaluation metrics}
We evaluate all estimators by measuring Spearman’s $\rho$ between the true and generated orderings of all instances in the test data. 
We obtain the generated ordering by sorting instances according to the predicted annotation time.
For our adaptive estimators that explicitly learn to predict the annotation time, we further report the mean absolute error (MAE), the rooted mean squared error (RMSE), and the coefficient of determination ($R^2$).

\paragraph{Models and features}
For an effective deployment in interactive annotation scenarios, we require models that are capable of fast training and inference.
We additionally consider the amount of computational resources that a model requires as they pose further limitations for the underlying annotation platform. 
Consequently, fine-tuning large language models such as BERT is infeasible as they require long training times and a large amount of computational resources.\footnote{Note that using such models would require an annotation platform to either deploy an own GPU or buy additional computational resources from external providers.}
Instead, we utilize a combination of neural embeddings obtained from a large pre-trained language model combined with an efficient statistical model.
As our goal is to predict the total time an annotator requires to annotate an instance (i.e., a sentence or a paragraph), we further require a means to aggregate token- or subtoken-level embeddings that are used in recent language models~\citep{sennrich-etal-2016-neural}.
One such solution is S-BERT~\citep{reimers-gurevych-2019-sentence} which has shown a high performance across various tasks. 
Moreover, \citet{reimers-gurevych-2019-sentence} provide S-BERT for a variety of BERT-based models, allowing future study conductors to easily extend our setup to other languages and specific tasks.
For computational efficiency, we use the \textit{paraphrase-distilroberta-base-v1} model which utilizes a smaller, distilled RoBERTa model~\citep{Sanh2019DistilBERTAD}. 
As a comparison to S-BERT, we further evaluate bag-of-words (BOW) features for all three models (cf.~\cref{tab:exp_hyper1}).
For the Ridge Regression (RR), Gaussian Process Regression (GP), and GBM Regression (GBM) models, we use the implementations of \citet{scikit-learn} and \citet{lightgbm2017}.

\begin{table*}[t]
	\centering
	\begin{tabular}{@{}llrrrrr@{}}
		\toprule
		Name                     & Features   &              MAE &             RMSE &                           $R^2$ &   $\rho$ &        t \\
		\midrule
		RR($\alpha=0.5$ )                    & BOW        &             1.85 &             2.96 &                         0.47 &  0.73 &     0.42 \\
		RR($\alpha=0.5$ )                    & S-BERT      &             1.92 &             2.84 &                         0.51 &  0.79 &     \textbf{0.04} \\
		RR($\alpha=1$ )                    & BOW        &             1.80 &             2.91 &                         0.49 &  0.74 &    0.41 \\
		RR($\alpha=1$ ) *                    & S-BERT      &             1.89 &             2.82 &                         0.52 &  0.79 &     0.04 \\
		GP(kernel=Dot + White) & BOW        &             1.82 &             2.93 &                         0.48 &  0.74 &  257.67 \\
		GP(kernel=Dot + White) * & S-BERT      &             \textbf{1.80} &             \textbf{2.76} &                         \textbf{0.54} &  \textbf{0.81}  &  14.35 \\
		GP(kernel=RBF(1.0) & BOW        &             5.33 &             6.71 &                        -1.73 & -0.12 &  300.38 \\
		GP(kernel=RBF(1.0) & S-BERT      &             5.33 &             6.71 &                        -1.73 & -0.12 &   32.66 \\
		GBM            & BOW        &             2.07 &             3.26 &                         0.36 &  0.68 &     0.25 \\
		GBM *            & S-BERT      &             1.83 &             2.83 &                         0.52 &  0.79 &     2.98 \\
		\bottomrule
	\end{tabular}
	\caption{Hyper-parameter tuning for adaptive estimators. We train on Muc7\textsubscript{T} A and evaluate on its development set.
		$t$ denotes the total time for training and prediction on the whole dataset. 
		Best parameters are marked by * and the best scores are highlighted in \textbf{bold}.
		We report the mean absolute error (MAE), the rooted mean squared error (RMSE), Spearman's $\rho$, and the coefficient of determination ($R^2$). 
	}
	\label{tab:exp_hyper1}
	
\end{table*}

\paragraph{Hyper-parameter tuning}
We use the full experimental setup to identify the best performing parameters for our experiments using simulated annotators.
We evaluate different values for regularization strength ($\alpha$) for RR and we evaluate different kernel functions for GP.
To ensure that the required training of our adaptive estimators does not negatively affect the annotations due to increased loading times and can be realistically performed during annotation, we further measure the overall training time (in seconds). 
We use the development split of Muc7\textsubscript{T} A to tune our hyper-parameters for all models used across all datasets.
Considering the small number of training instances in both datasets, we do not tune SigIE- or SPEC-specific hyperparameters.  
All experiments were conducted using an \textit{AMD Ryzen 5 3600}.
\cref{tab:exp_hyper1} shows the results of our hyper-parameter tuning experiments.
Overall, we find that S-BERT consistently outperforms BOW in terms of Spearman's $\rho$.
As the result of the hyper-parameter tuning, we use S-BERT embeddings as input features and evaluate GP with a combined dot- and white-noise kernel and RR with $\alpha=1$ in our adaptive experiments.

\begin{table*}[t]
	\centering
	\begin{tabular}{@{}ccccccc@{}}
		\toprule
		Name    & Model                    &   MAE &   RMSE &    $R^2$ &   $\rho$ &    t \\
		\midrule
		& RR                    &  1.87 &   2.68 &  0.56 &  0.80 & 0.15 \\
		Muc7\textsubscript{T} A & GP &  1.79 &   2.66 &  0.57 &  0.82 & 7.23 \\
		& GBM            &  1.95 &   2.97 &  0.47 &  0.75 & 3.40 \\ \midrule
		
		& RR                    &  2.19 &   3.42 &  0.44 &  0.79 & 0.02 \\
		Muc7\textsubscript{T} B & GP  &  2.08 &   3.37 &  0.46 &  0.81 & 8.85 \\
		& LGBM            &  2.13 &   3.50 &  0.41 &  0.75 & 2.90 \\ \midrule
		& RR                    &  7.96 &   9.50 &  0.46 &  0.73 & 0.00 \\
		SigIE   & GP &  7.62 &   9.60 &  0.44 &  0.70 & 0.08 \\
		& GBM            &  8.22 &  10.84 &  0.29 &  0.55 & 0.14 \\ \midrule
		& RR                    &  9.63 &  13.86 & -0.14 &  0.50 & 0.03 \\
		SPEC    & GP &  7.63 &  12.07 &  0.14 &  0.51 & 0.73 \\
		& GBM            &  8.05 &  12.50 &  0.07 &  0.35 & 1.70 \\
		\bottomrule
	\end{tabular}
	\caption{Performance of the best performing adaptive estimators on the four datasets (Muc7\textsubscript{T} provides annotation times from two different annotators A and B) trained on the respective train and evaluated on their test splits. 
		We report the mean absolute error (MAE), the rooted mean squared error (RMSE), the coefficient of determination ($R^2$) and Spearman's $\rho$. 
	}
	\label{tab:exp_regression_full}
	
\end{table*}

\subsection{Experimental Results}
We first report our experimental results for the full and adaptive setup. 
For conducting our experiments with simulated annotators, we use the best performing models from our hyper-parameter tuning of the respective models on the Muc7\textsubscript{T} dataset and report the results of the best performing models.

\paragraph{Full results}
\cref{tab:heuristics-test} shows the results for the heuristic estimators and regression models evaluated on the test split of each dataset.
We find that heuristics that mainly consider length-based features ($\mathrm{sen}$ and $\mathrm{FK}$) are not suited for the SigIE data that consists of email signatures.
One reason for this may be the different text type of email signatures in comparison to Newswire articles and PubMed abstracts. 
More specifically, analyzing the ratio between non-alphabetical or numeric characters (excluding @ and . ) and other characters shows that SigIE contains a substantial amount of characters that are used for visually enhancing the signature (some are even used in text art).
Overall, 29.9\% of the characters in SigIE are non-alphabetical or numeric in contrast to 16.7\% in SPEC and 19.9\% in Muc7\textsubscript{T}.\footnote{The Twitter data we introduce in \cref{user_study} consists of 20.7\% non-alphabetical or numeric characters.}
Considering that only 1.7\% of them appear within named entities in SigIE (such as + in phone numbers) most of them rather introduce noise especially for length based-features such as $\mathrm{sen}$ and $\mathrm{FK}$. 
On Muc7\textsubscript{T} and \textsc{SPEC}, all three heuristics produce an ordering that correlates with annotation time to some extent.
On average, $\mathrm{mlm}$ is the best performing and most robust heuristic across all three datasets.
For our adaptive estimators, RR and GP both similarly outperform GBM in terms of Spearman's $\rho$.
However, we can find that GP consistently outperforms RR and GBM in terms of MAE and RMSE, as well as in terms of $R^2$ on Muc7\textsubscript{T} and SPEC. 
We report the extensive results in \cref{tab:exp_regression_full}.

\begin{table}[thb]
	\centering
	\begin{tabular}{@{}lcccccc@{}}
		\toprule
		Dataset   &   $\mathrm{sen}$ &   $\mathrm{FK}$ & $\mathrm{mlm}$ & $\mathrm{RR}$ &  $\mathrm{GP}$ & $\mathrm{GBM}$ \\        \midrule
		Muc7\textsubscript{T} A    &   0.60 &  0.37 &  0.57 & 0.80 & \textbf{0.82}  & 0.75 \\
		Muc7\textsubscript{T} B     &   0.60 &  0.38 &  0.55 & 0.79 & \textbf{0.81}  & 0.75 \\
		SigIE     &   0.08 &  0.01 &  0.59  & \textbf{0.73} & 0.70 & 0.55 \\
		SPEC      &   \textbf{0.63} &  0.38 &  0.32 & 0.50 & 0.51 & 0.35 \\
		\midrule 
		Average & 0.48 & 0.29 & 0.52 & 0.71 & 0.71 & 0.60 \\
		\bottomrule
	\end{tabular}
	\caption{Spearman's $\rho$ between test data and the orderings generated by the evaluated heuristics  and adaptive models.} 
	\label{tab:heuristics-test}
\end{table}

\begin{figure*}[b]
	\begin{center}
		\subfloat[Muc7\textsubscript{T} A]{\includegraphics[width=.98\textwidth]{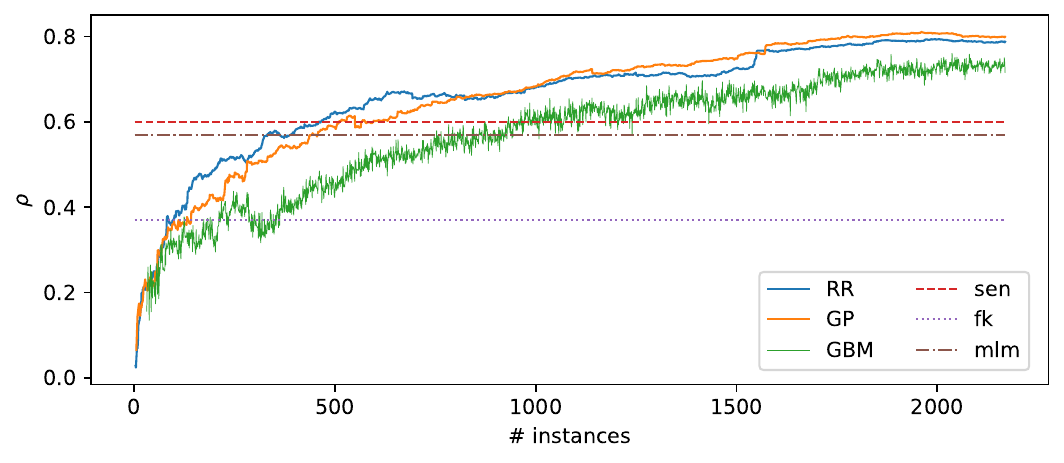}}  \newline 
		\subfloat[SigIE]{\includegraphics[width=.98\textwidth]{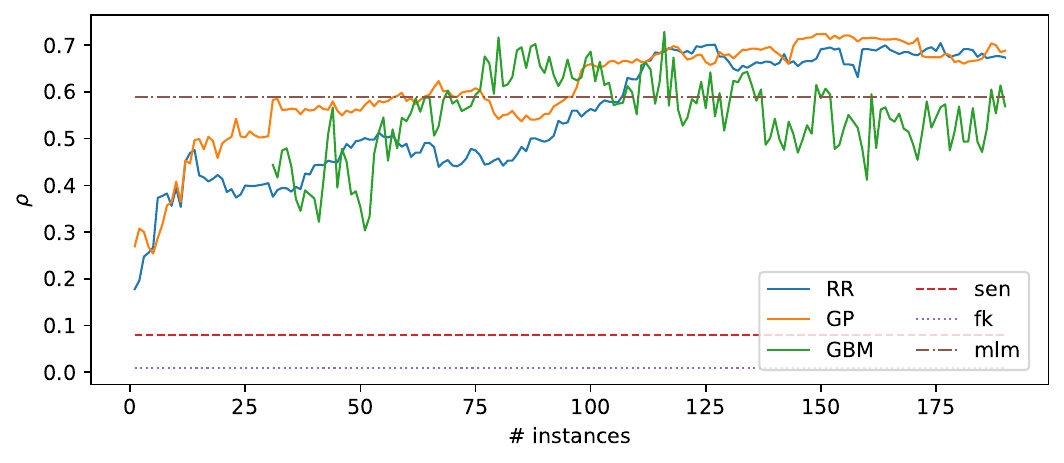}} \newline
		\subfloat[SPEC]{\includegraphics[width=.98\textwidth]{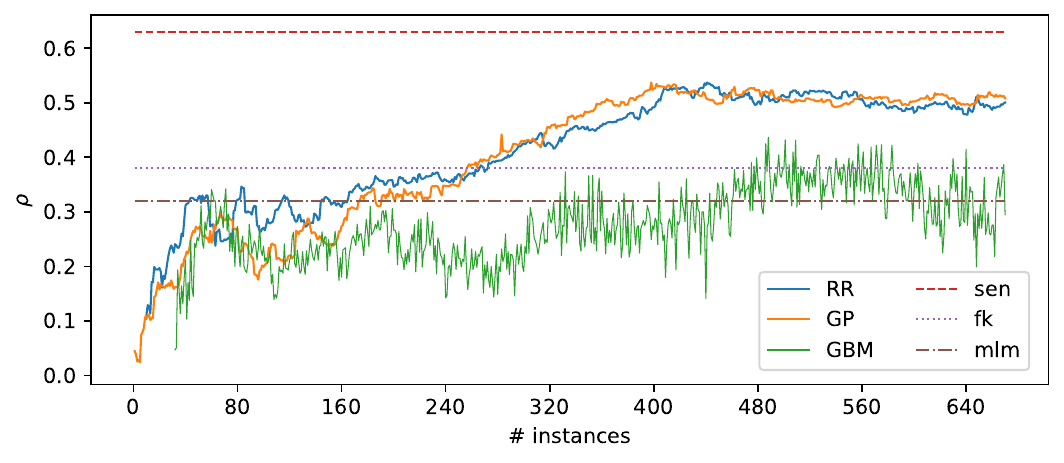}}  \newline    \end{center}
	\caption{Experimental results of our adaptive estimators with simulated annotators. Horizontal lines show the performance of the respective non-adaptive estimators. }
	\label{fig:experiments_simulation}
\end{figure*}

\paragraph{Adaptive results}
To evaluate the performance of adaptive estimators with increasing numbers of annotated instances, we perform experiments with simulated annotators.
At each iteration, we use a model trained on the already annotated data to select the instance with the lowest predicted annotation time (randomly in the first iteration). 
The simulated annotator then provides the respective gold annotation time which is then added to the training set. 
Finally, the model is re-trained an evaluated on the test data.
These steps are repeated until all instances are annotated.
\cref{fig:experiments_simulation} shows the  Spearman's $\rho$ performance of all three models after each iteration across all datasets. 
We can observe that all models display a rather steep learning curve after training on only a few examples, despite suffering from a cold start in early iterations.
Moreover, we find that GP and RR are capable of outperforming $\mathrm{mlm}$ consistently after 100-500 instances.
GBM shows the weakest performance and is consistently outperformed by the other models for Muc7\textsubscript{T} and SPEC while being rather noisy. 
Although we find that non-adaptive estimators can suffice especially in early iterations, our experiments also show the potential of adaptive estimators with an increasing number of annotations.
This indicates that hybrid approaches that combine non-adaptive and adaptive estimators could be an interesting direction for future work.
For instance, one may consider using non-adaptive estimators in early stages until a sufficient amount of annotated instances are available to train more reliable adaptive estimators.
Another approach could be to combine the rankings of different estimators, for instance, via Borda count \citep{szpiro2010numbers} or learn a weighting of the individual estimators.

\clearpage
\section{Human Evaluation}
\label{user_study}
To evaluate the effectiveness of our easy-instances-first AC with real annotators, we conduct a user study on a classification task for English tweets and analyze the resulting annotations in terms of \textit{annotation time} and \textit{annotation quality}.
We design the study to not require domain-expertise and conduct it with citizen science volunteers.\footnote{We provide a statement regarding the conduct of ethical research after the conclusion.} 

\paragraph{Hypothesis}
We investigate the following hypothesis: Annotators that are presented with easy instances first and then with instances that gradually increase in terms of annotation difficulty require less annotation time or have improved annotation quality compared to annotators that receive the same instances in a random order.

\subsection{Study Design}
A careful task and data selection is essential to evaluate AC, as our goal is to measure differences that solely result from a different ordering of annotated instances. 
We also require instances with varying difficulty, further restricting our study design in terms of task and data.

\paragraph{Data source}
To avoid compromising the study results due to noisy data, we use an existing corpus that has been carefully curated and provides gold labels for evaluating the annotation quality.
To involve non-expert annotators, we furthermore require data that does not target narrow domains or requires expert knowledge.
As such, tasks such as identifying part-of-speech tags would substantially reduce the number of possible study participants due to the required linguistic knowledge.
We identify COVIDLies~\citep{hossain-etal-2020-covidlies} as a suitable corpus due to the current relevance and the high media-coverage of the Covid-19 pandemic; ensuring a sufficient number of participants that are well-versed with the topic.
The corpus consists of English tweets that have been annotated by medical experts with one out of 86 common misconceptions about the Covid-19 pandemic.
Each instance consists of a tweet-misconception pair and if the tweet "agrees", "disagrees", or has "no stance" towards the presented misconception. 

\paragraph{Annotation task}
Using the COVIDLies corpus as our basis, we define a similar task that is better suited for lay people and that allows us to explicitly control the annotation difficulty.
We restrict the task to identifying the most appropriate misconception out of six possible choices.
We furthermore only include tweets that agree with a misconception (i.e., we do not ask for a stance annotation) to avoid interdependencies between stance and misconception annotations that may introduce additional noise to the results and put an unnecessary burden on the participants.\footnote{We experimented with including stance annotations (positive, negative, or neutral) during early stages of our study setup but removed them due to a substantially increased overall annotation difficulty.}
To exclude further sources of noise for our study, we manually check all tweets and remove all duplicates (possibly due to retweets) and hyperlinks to increase readability and avoid distractions.
We furthermore remove all tweets that were malformed (i.e., ungrammatical or containing several line breaks) or linked to misconceptions with less than five semantically similar candidates that could serve as distractors.\footnote{The sets of similar misconceptions were manually created as explained in the next paragraph.}
For the final selection, we choose the 60 shortest tweets.

\begin{figure*}[htb]
	\centering
	\subfloat[Easy Example]{
		\includegraphics[width=0.85\textwidth]{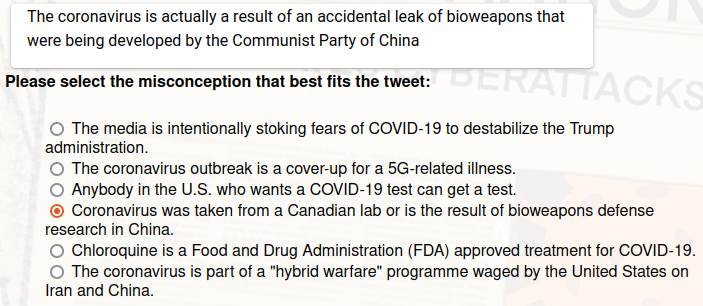}}
	
	\subfloat[Difficult Example]{
		\includegraphics[width=0.85\textwidth]{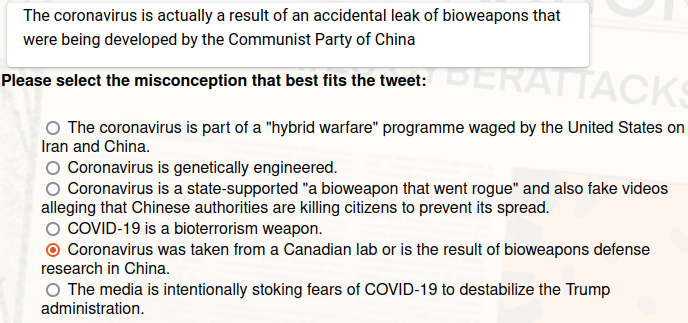}}
	\caption{Example tweet from the user study with an easy misconception set (used in the study) and a difficult misconception set.}
	\label{fig:example-user-study}
	
\end{figure*}

\paragraph{Distractor selection}
The goal of the study is to observe effects that solely result from the ordering of instances with varying annotation difficulty. 
Hence, we need to ensure that annotated instances correspond to specific difficulties and are balanced equally for each participant.
To control the annotation difficulty, we construct five possible sets of misconceptions for each instance that are presented to the annotator; each corresponding to a respective difficulty-level ranging from "very easy" to "very difficult".
Each set consists of the expert-selected misconception and five additional misconceptions that serve as distractors which are commonly used in cloze-tests~\citep{taylor1953cloze}.
Following existing research on automated cloze-test generation, we focus on \textbf{semantic similarity} to generate distractor subsets~\citep{agarwal-mannem-2011-automatic,mostow2012generating,yeung-etal-2019-difficulty} and manually create one set of five semantically dissimilar and one set of five semantically similar misconceptions for each misconception.\footnote{Initially, we also investigated the use of recent automated approaches to create those subsets~\citep{gao-etal-2020-distractor}. 
	However, the resulting subsets rather targeted syntactic instead of semantic similarity. 
	One reason for this may be that approaches to generate cloze-tests consider only single-token gaps whereas the misconceptions consist of several words that form a descriptive statement.}
As semantically dissimilar distractors are much easier to identify than semantically similar ones~\citep{mostow2012generating}, we can manipulate annotation difficulty by adapting the number of semantically similar distractors; that is, starting from the set of dissimilar (very easy) misconceptions, we can gradually increase the difficulty by replacing a dissimilar misconception with a similar one until only the set of similar (very difficult) misconceptions remains.
\Cref{fig:example-user-study} shows a tweet from our user study with its respective easy and difficult misconception sets.
As can be seen, the difficult misconception set consists of two more semantically similar misconceptions.
Especially notable is the third misconception that states the opposite of the tweet's misconception but with a similar wording. 

\subsection{Study Setup}
We set up our evaluation study as a self-hosted web application that is only accessible during the study (one week). 
Participants can anonymously participate with a self-chosen, unique study key that allows them to request the deletion of their provided data at a later point.
Upon registration, they are informed about the data presented and collected in the study, its further use, and the purpose of the study.
Before collecting any data, participants are explicitly asked for their informed consent.
Overall, we recruited 40 volunteers that provided their informed consent to participate in our study and annotated 60 instances each.

\paragraph{Participants}
Our volunteers come from a variety of university majors, native languages, English proficiency and annotation experience backgrounds.
All participants provided a rather high self-assessment of English proficiency, with the lowest proficiency being intermediate (B1) provided by only one participant.
70\% of the participants stated an English proficiency-level of advanced (C1) or proficient (C2).
Most participants have a higher level of education and are university graduates with either a Bachelor's or Master's degree; however, none of them have a medical background which may have given them an advantage during the annotation study. 
Upon completing the annotations, all participants received a questionnaire including general questions about their previous annotation experience and perceived difficulty of the task (cf. \cref{sec:questionnaire}).

\paragraph{Ordering strategy}
All participants are randomly assigned to one out of four groups (ten participants per group), each corresponding to a strategy that leads to a different ordering of annotated instances.
We investigate the following strategies:
\begin{description}[noitemsep,topsep=3pt,itemsep=3pt,itemindent=-1em]
	\item[Random] is the control group that consists of randomly ordered instances.
	\item[AC$_\mathrm{mlm}$] uses the masked language modeling loss.
	It is a pre-computed, heuristic estimator and had (on average) the highest and most stable correlation to annotation time in our experiments with simulated annotators.
	\item[AC$_\mathrm{GP}$] uses a Gaussian Process that showed the highest performance on the sentence-labeling task (SPEC) in our simulated annotator experiments (cf. \cref{tab:heuristics-test}). It is trained interactively to predict the annotation time.
	We train a personalized model for each annotator using S-BERT embeddings of the presented tweet.
	\item[AC$\mathrm{gold}$] consists of instances explicitly ordered from very easy to very difficult using the pre-defined distractor sets. 
	Although such annotation difficulties are unavailable in real-world annotation studies, it provides an upper-bound for the study.
\end{description}

\paragraph{Control instances}
To provide a fair comparison between different groups, we further require participants to annotate instances that quantify the difference with respect to prior knowledge and annotation proficiency.
For this, we select the first ten instances and present them in the same order for all annotators.
To avoid interdependency effects between the control instances and the instances used to evaluate AC$_{\{*\}}$, we selected instances that have disjoint sets of misconceptions.

\paragraph{Balancing annotation difficulty} 
We generate instances of different annotation difficulties using the sets of semantically similar and dissimilar misconceptions that serve as our distractors. 
We randomly assign an equal amount of tweet-misconception pairs to each difficulty-level ranging from very easy to very difficult.
The resulting 50 instances for our final study span similar ranges in terms of length as shown in \cref{fig:difficulty-char} which is crucial to minimize the influence of reading time on our results.
Overall, each of the five difficulty-levels consists of ten (two for the control instances) unique tweets that are annotated by all participants in different order.

\begin{table}[htb]
	\centering
	\begin{tabular}{@{}lccccc@{}}
		\toprule 
		\# Chars & very easy & easy & medium & difficult & very difficult \\
		\midrule
		T &  219 & 211 & 183 & 217 & 194 \\
		T \& MC & 638 & 603 & 599 & 586 & 593 \\
		\bottomrule
	\end{tabular}
	\caption{Average number of characters per tweet (T) and tweet and misconception (T \& MC) across all difficulty-levels of annotated items.}
	\label{fig:difficulty-char}
\end{table}

\paragraph{Study process}
The final study consists of 50 instances that are ordered corresponding to the group a participant has been assigned to.
Each instance consists of a tweet and six possible misconceptions (one expert-annotated and five distractors) from which the participants are asked to select the most appropriate one. 
The lists of the six presented misconceptions are ordered randomly to prevent that participants learn to annotate a specific position.
Finally, we ask each participant to answer a questionnaire that measures the perceived difficulty of the annotated instances.

\subsection{General Results}
In total, each of the 40 participants has provided 60 annotations, resulting in 400 annotations for the ten control instances (100 per group) and 2,000 annotations for the 50 final study instances (500 per group). 
In terms of annotation difficulty, each of the five difficulty-levels consists of 80 annotations for the control instances and 400 annotations for the final study.
To assess the validity of AC$_{\{*\}}$, we require two criteria to be fulfilled:
\begin{itemize}[topsep=5pt,itemsep=3pt]
	\item[\textbf{H1}] The participant groups do not significantly differ in terms of annotation time or annotation quality for the control instances.
	\item[\textbf{H2}] AC$_{\{*\}}$ shows a significant difference in annotation time or annotation quality compared to Random or each other.
\end{itemize}

\paragraph{Outliers}
Across all 2,400 annotations, we identify only two cases where participants required more than ten minutes for annotation and are apparent outliers.
To avoid removing annotations for evaluation, we compute the mean and standard deviation of the annotation time across all annotations (excluding the two outliers) and set the maximum value to $t_{\max}=\mu + 5 \sigma = 156.39$ seconds. 
This results in ten annotations that are set to $t_{\max}$ for Random, three for AC$_{\mathrm{mlm}}$, one for AC$_{\mathrm{GP}}$, and zero for AC$_{\mathrm{gold}}$.
Note that this mainly favors the random control group that serves as our baseline.

\begin{table}[htb]
	\centering
	\begin{tabular}{@{}lcccccc@{}}
		\toprule
		&  $\Sigma_t$ & $\mu_t$ &   $\sigma_t$ &  25\% &   50\% &   75\% \\
		\midrule
		Random   &  1,852.9 &  27.3 &  27.2 &  12.9 &  18.2 &  29.5 \\
		AC$_{\mathrm{mlm}}$  & 1,273.4  & 23.2 &  19.4 & \textbf{11.7} &  18.6 &  27.4 \\
		AC$_{\mathrm{GP}}$ & 1,324.3 & 26.4 &  19.0 &  14.9 &  20.7 &  30.8 \\
		AC$_{\mathrm{gold}}$  & \textbf{1,059.6} & \textbf{21.2} &  \textbf{12.8} &  12.6 &  \textbf{18.0} &  \textbf{26.5} \\
		\bottomrule
	\end{tabular}
	\caption{Mean, standard deviation, and 25\%, 50\%, and 75\% percentiles of annotation (in seconds). $\Sigma_t$ denotes the total annotation time an annotator of the respective group required to finish the study (on average).} 
	\label{tab:annotation-time}
\end{table}

\paragraph{Annotation time}
\cref{tab:annotation-time} shows the results of the final study in terms of annotation time per group.
Overall, annotators of AC$_{\mathrm{gold}}$ required on average the least amount of time per instance and had the lowest standard deviation.
We also observe a substantial decrease in the maximum annotation time, as shown in the 75th percentile for AC$_{\mathrm{gold}}$.
Conducting a Kruskal--Wallis test~\citep{kruskal1952use} on the control instances across all participant groups results in a p-value of $p=0.200 < 0.05$.\footnote{In general, ANOVA (analysis of variance) is a more expressive test that does not require pairwise comparisons that are necessary for the less expressive Kruskal--Wallis test.
	However, we cannot apply ANOVA in our case due to violated conditions on normality and homoscedasticity of the collected data.}
Hence, we cannot reject the null-hypothesis for the control instances and conclude, that all groups initially do not show statistically significant differences in terms of annotation time for the control instances, thereby satisfying H1.
Next, we conduct the same test on the evaluation instances and observe a statistically significant p-value of $p=4.53^{-6} < 0.05$.
For a more specific comparison, we further conduct pairwise Welch's t-test~\citep{welch1951comparison} for each strategy with a Bonferroni-corrected p-value of $p=\frac{0.05}{6}=0.008\overline{3}$ to account for multiple comparisons~\citep{bonferroni1936teoria}.
Overall, AC$_{\mathrm{gold}}$ performs best, satisfying H2 with statistically significant improvements over Random ($p=7.28^{-6}$) and \mbox{AC$_{\mathrm{GP}}$} ($p=3.79^{-7}$).
Although the difference to AC$_{\mathrm{mlm}}$ is substantial, it is not statistically significant ($p=0.0502$).
The best performing estimator is AC$_{\mathrm{mlm}}$ that performs significantly better than Random ($p=0.0069$) and substantially better than AC$_{\mathrm{GP}}$ ($p=0.0084$).
Between AC$_{\mathrm{GP}}$ and Random, we cannot observe any statistically significant differences ($p=0.5694$).

\begin{table}[htb]
	\centering
	\begin{tabular}{@{}lccccc@{}}
		\toprule
		&   $\mu_{acc}$ &   $\sigma_{acc}$ &  25\% &   50\% &   75\% \\
		\midrule
		Random                        & 84.7 &  4.22 &  82.0 &  \textbf{86.0} & \textbf{ 88.0} \\
		AC$_{\mathrm{mlm}}$  & 83.6 &  5.32 & 80.0 &  84.0 &  86.0 \\
		AC$_{\mathrm{GP}}$   & 83.6 &  \textbf{2.95} &  82.0 &  \textbf{86.0} &  86.0 \\
		AC$_{\mathrm{gold}}$ & \textbf{85.6} &  3.01 &  \textbf{84.0} &  84.0 &  \textbf{88.0} \\
		\bottomrule
	\end{tabular}
	\caption{Mean, standard deviation, and 25\%, 50\%, and 75\% percentiles of annotation quality (in percent accuracy).}
	\label{tab:annotation-acc}
\end{table}

\paragraph{Annotation quality}
We evaluate annotation quality by computing the accuracy for each participant, that is, the percentage of misconceptions that they were able to correctly identify out of the six presented ones.
\cref{tab:annotation-acc} shows our results in terms of accuracy.
Although AC$_{\mathrm{gold}}$ has the highest mean accuracy, the most differences lie within the range of $2$\% accuracy which is equivalent to only a single wrongly annotated instance.
Conducting Kruskal--Wallis tests for the control instances shows that the difference in terms of accuracy is not statistically significant ($p=0.881$), satisfying H1.
However, the same test shows no statistically significant difference for the final study ($p=0.723$). 
One reason for this may be our decision to conduct the study with voluntary participants and their higher intrinsic motivation to focus on annotation quality over annotation time~\citep{chau-etal-2020-understanding}. 
In contrast to crowdsourcing scenarios where annotators are mainly motivated by monetary gain --- trying to reduce the amount of time they spend on their annotation at the cost of quality --- voluntary annotators are more motivated to invest additional time to provide correct annotations; even more so in a setup with a low number of 60 instances.

\vspace{1em}

\begin{figure*}[thb]
	\begin{minipage}{0.48\textwidth}
		\raggedleft
		\includegraphics[width=1.0\textwidth]{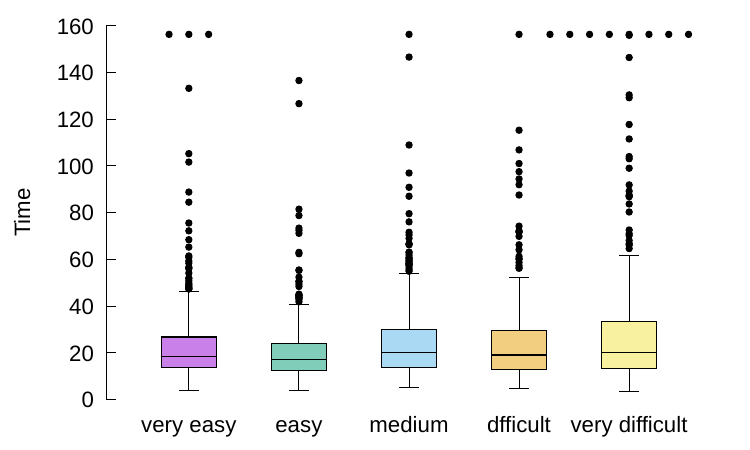}
		\caption{Annotation time (in seconds) grouped by difficulty level.}
		\label{fig:difficulty-times}
	\end{minipage}\hfill 
	\begin{minipage}{0.48\textwidth}
		\raggedright
		\includegraphics[width=1.0\textwidth]{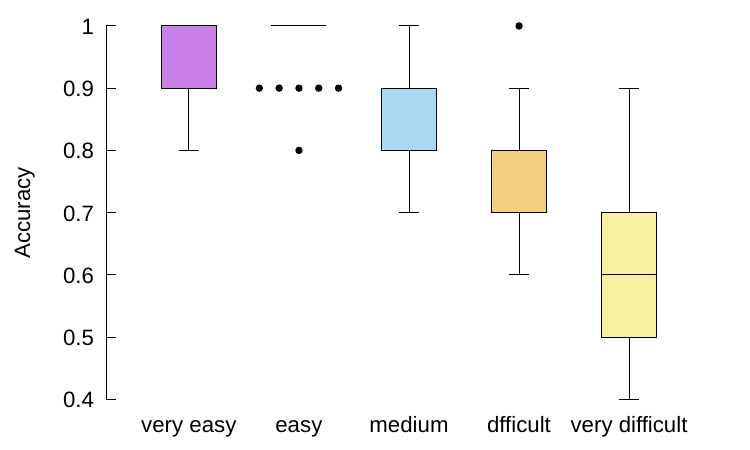}
		\caption{Accuracy per annotator grouped by difficulty level.} 
		\label{fig:difficulty-acc}
	\end{minipage}
\end{figure*}

\begin{figure}
	\centering
	\includegraphics[width=0.59\textwidth]{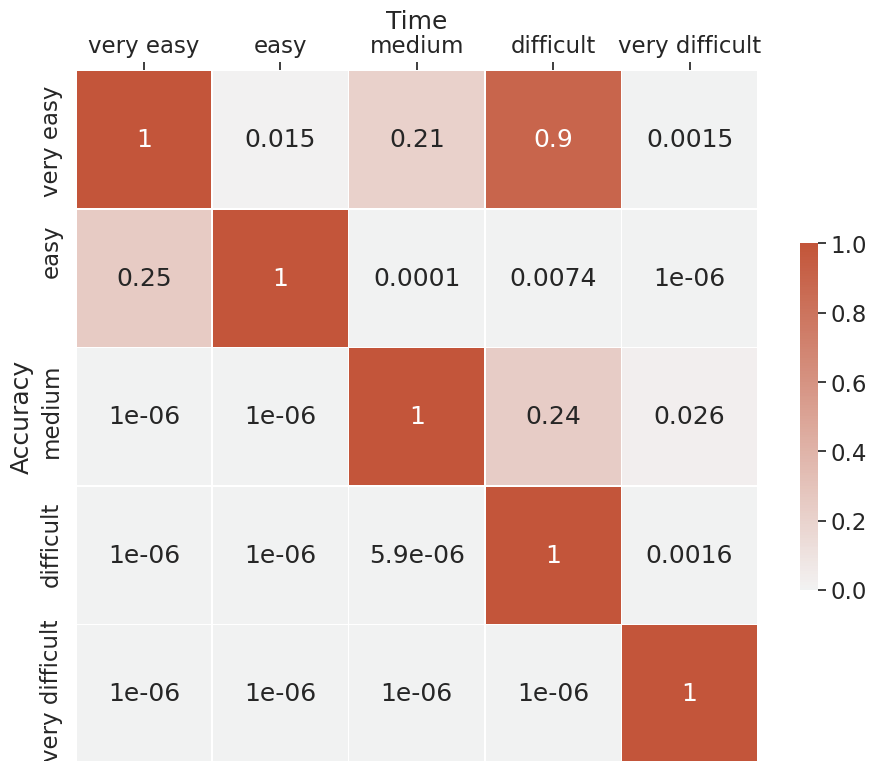}
	\caption{P-values for time (in seconds) and accuracy between different difficulty levels.}
	\label{fig:pvals-difficulty}
\end{figure}

\paragraph{Difficulty evaluation}
To validate our generation approach with distractors, we further evaluate all annotation instances in terms of their annotation difficulty.
As \cref{fig:difficulty-times} and \cref{fig:difficulty-acc} show, one can observe non-negligible differences in terms of annotation time as well as accuracy across instances of different difficulties.
Conducting pairwise Welch's t-tests with a Bonferroni corrected p-value of $p=\frac{0.05}{10}=0.005$ shows that in terms of accuracy, only very easy and easy instances do not express a statistically significant difference ($p=0.25$), showing that participants had more trouble in identifying the correct misconception for difficult instances.\footnote{Overall, we require $\frac{n(n-1)}{2}$ pairwise comparisons resulting in 10 comparisons with $n=5$.}
For all other instances, we observe p-values smaller than 1e$^{-6}$ as shown in \cref{fig:pvals-difficulty}.
In terms of annotation time, the differences are not as apparent as in annotation accuracy. 
We find statistically significant differences in only four out of ten cases showing that the annotation difficulty does not necessarily impact the annotation time.
Overall, we still observe that instances express significant differences in terms of either annotation time or quality (or both), showing that our approach using distractor sets to control the annotation difficulty worked well.

\clearpage

\subsection{Error Analysis}\label{sec:error-analysis}

\begin{table}[h!tb]
	\centering
	\begin{tabular}{@{}lcccccc@{}}
		\toprule
		&   $\mu_{t}$ &   $\sigma_{t}$ &  25\%  &  50\%  &  75\%  \\ 
		\midrule
		MAE  &  12.4  &  6.1  &    8.5  & 10.4  & 14.3  \\ 
		RMSE &  17.2  &  9.1  &  11.1  & 13.9  & 20.3  \\ 
		$R^2$   &  0.0  &  0.0  &  -0.1  & 0.0  & 0.0  \\ 
		$\rho$  &  -0.1  &  0.2  & -0.3  & -0.1  &  0.1  \\ 
		\bottomrule
	\end{tabular}
	\caption{Leave-one-out cross validation results on annotation times, grouped by user and averaged.}
	\label{tab:ac_at_error}
\end{table}

\begin{figure}[htb]
	\begin{minipage}{0.49\textwidth}
		\centering
		\includegraphics[width=1.0\textwidth]{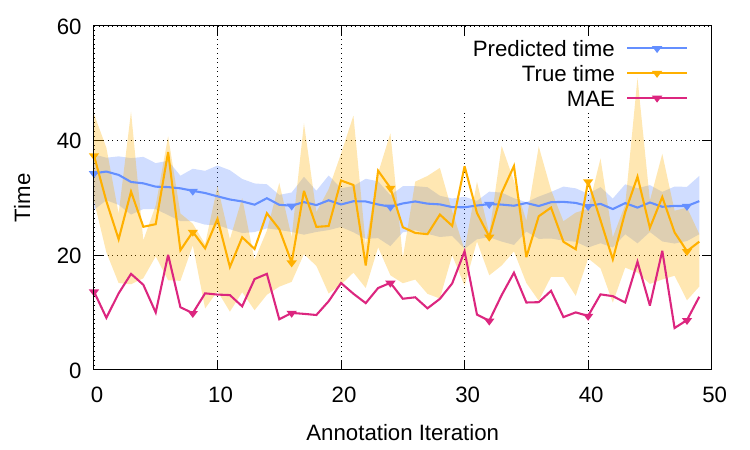}
		\caption{Mean, lower, and upper percentiles for predicted and true annotation time and the mean absolute error.}
		\label{fig:interactive-ordering}
	\end{minipage}\hfill 
	\begin{minipage}{0.49\textwidth}
		\centering
		\includegraphics[width=1.0\textwidth]{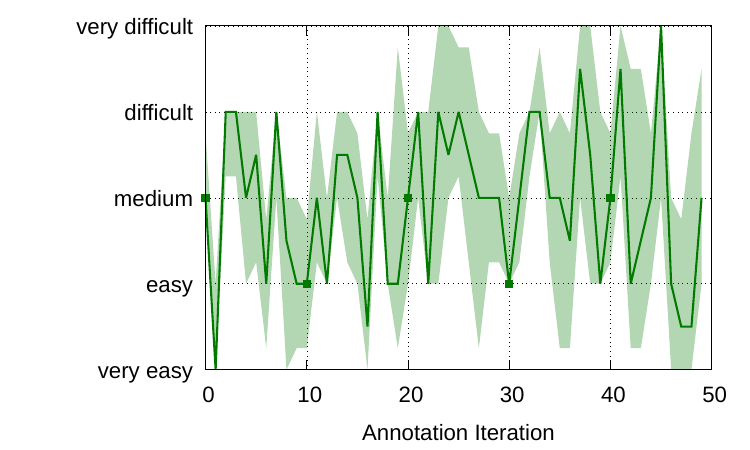}
		\caption{Median, lower, and upper percentile for instance difficulty with AC$_{\mathrm{GP}}$ at each iteration.}
		\label{fig:gp-difficulty}
	\end{minipage}   
\end{figure}

\paragraph{Model performance} While AC$_{\mathrm{mlm}}$ and AC$_{\mathrm{gold}}$ both outperform the random baseline significantly, AC$_{\mathrm{GP}}$ does not.
To analyze how well the used GP model performs for individual annotators, we perform leave-one-user-out cross validation experiments across all 40 participants.
\cref{tab:ac_at_error} shows the mean absolute error (MAE), the rooted mean squared error (RMSE), the coefficient of determination ($R^2$), and Spearman's $\rho$ of our experiments.
Overall, we find a low correlation between the predicted and true annotation time and high standard deviations across both errors. 
Further analyzing the performance of AC$_{\mathrm{GP}}$ for interactively predicting the annotation time (cf.~\cref{fig:interactive-ordering}) shows that the model adapts rather slowly to additional data.
As can be observed, the low performance of the model (MAE between $10-20$ seconds) results in a high variation in the annotation time of the selected instances between subsequent iterations; further experiments strongly suggest this is due to the model suffering from a cold start and the small amount of available training data as also discussed below.

\paragraph{Correlation with AC$_{\mathrm{gold}}$}
A second shortcoming of AC$_{\mathrm{GP}}$ becomes apparent when observing the difficulty of the sampled instances across all iterations, shown in \cref{fig:gp-difficulty}.
We observe a low Spearman's $\rho$ correlation to AC$_{\mathrm{gold}}$ of 0.005, in contrast to AC$_{\mathrm{mlm}}$ ($\rho=0.22$). 
Only Random has a lower correlation of $\rho=-0.15$. 
This shows that model adaptivity plays an important role especially in low-data scenarios such as in early stages during annotation studies. 
We plan to tackle this issue in future work using more sophisticated models and combined approaches that initially utilize heuristics and switch to interactively trained models with the availability of sufficient training data.

\subsection{Participant Questionnaire}\label{sec:questionnaire}
After completing the annotation study, each participant answered a questionnaire quantifying their language proficiency, previous annotation experience, and perceived difficulty of the annotation task.

\paragraph{Language proficiency}
In addition to their CEFR language proficiency~\citep{council2001common}, we further asked participants to provide optional information about their first language and the number of years they have been actively speaking English.
On average, our participants have been actively speaking English for more than 10 years.
Overall, they stated a language proficiency of: B1 (1), B2 (11), C1 (17), and C2 (11).
Most of our participants stated German as their first language (30). 
Other first languages include Vietnamese (4), Chinese (3), Russian (1), and Czech (1).\footnote{One participant decided not to disclose any additional information except their English proficiency.}

\paragraph{Annotation experience} 
We further collected data from our participants regarding their previous experience as study participants as well as study conductors.
In general, about 50\% of our participants (18) had not participated in annotation studies before. 
19 had participated in a few (one to three) studies, and only three in more than three studies.
Even more participants had not previously conducted a study (24) or only a few (12).
In total, four participants stated that they had setup more than three annotation studies.

\begin{table}[h!tb]
	\centering
	\begin{tabular}{@{}lcccccc@{}}
		\toprule
		&  \multicolumn{2}{c}{CEFR} & \multicolumn{2}{c}{Annotator} &  \multicolumn{2}{c}{Conductor} \\ 
		& $\rho$ & p-value & $\rho$ & p-value & $\rho$ & p-value \\
		\midrule
		Time  &  -0.307  &  0.054 & -0.134  & 0.409  & 0.085 & 0.600  \\ 
		Accuracy &  0.319  &  0.044 & -0.060  & 0.711  & -0.211 & 0.191 \\ 
		\bottomrule
	\end{tabular}
	\caption{Spearman's $\rho$ correlation analysis for three potential confounding factors.}
	\label{tab:confounders}
\end{table}

\paragraph{Confounding factors}
We identify the language proficiency and previous experience with annotation studies as potential confounding factors~\citep{vanderweele2013definition}.
Confounding factors are variables that are difficult to control for, but have an influence on the whole study and can lead to a misinterpretation of the results.
Especially in studies that include a randomized setup such as in ours --- due to the random assignment of our participants into the four groups --- it is crucial to investigate the influence of potential confounding factors.
In our analysis, we focus on variables for which all participants provided an answer, namely, their CEFR level and their experience as participants in and conductors of annotation studies (some of our participants were researchers).
\cref{tab:confounders} shows the results of a Spearman's $\rho$ correlation analysis for all three variables against annotation time and accuracy.
As we can see, the participants' experiences as annotators (Annotator) or study conductors (Conductor) only yields a low, non-significant correlation with time and accuracy and consequently, can be excluded as confounding factors.
The influence of their language proficiency (CEFR) is more interesting, as it shows a small negative correlation for annotation time and a small positive correlation for annotation accuracy with p-values around $0.05$, meaning that participants with a lower CEFR level required less time, but also had a lower accuracy. 
To investigate the influence of a participant's language proficiency on our results, we conduct a Kruskal--Wallis test for the distribution of different language proficiency levels across the four groups and find that they do not differ significantly with a p-value of $p=0.961$.
Nonetheless, we find that the CEFR level is an important confounding factor that needs to be considered in future study setups.

\paragraph{Perceived difficulty}
To quantify if there exists any difference between the actual difficulty and the perceived difficulty, we further asked our participants the following questions: 
\begin{itemize}
	\item[\textbf{PQ1:}] How difficult did you find the overall annotation task?
	\item[\textbf{PQ2:}] Did you notice any differences in difficulty between individual tweets?
	\item[\textbf{PQ3:}] Would you have preferred a different ordering of the items? 
\end{itemize}

\begin{figure}[htb]
	\centering
	\includegraphics[width=0.6\textwidth]{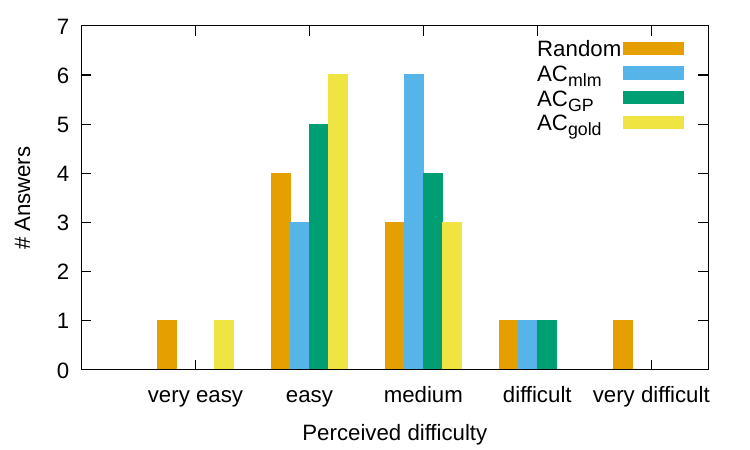}
	\caption{Accumulated perceived difficulty answers across all groups.}
	\label{fig:perceived-difficulty}
\end{figure}

\cref{fig:perceived-difficulty} shows the distribution of answers (from very easy to very difficult) to PQ1 across all four groups. 
Interestingly, whereas participants of the AC$_\mathrm{mlm}$ group did require less time during their annotation compared to AC$_\mathrm{GP}$, more people rated the study as of medium difficulty than participants of AC$_\mathrm{GP}$.
This may be an indicator that AC$_\mathrm{GP}$ may --- although not measurable in terms of annotation time --- alleviate the perceived difficulty for participants, hence, still reducing the cognitive burden.
We will investigate this in further studies that also include an item specific difficulty annotation, that is, by explicitly asking annotators for the perceived difficulty.\footnote{We excluded this additional annotation in the study as one pass already required $\sim 45-60$ minutes.}
Overall, only four out of 40 participants (two for AC$_\mathrm{GP}$ and one for AC$_\mathrm{mlm}$ and AC$_\mathrm{gold}$ each) did state to not have noticed any differences in terms of difficulty between different instances; showing that the selected distractors resulted in instances of noticeably different annotation difficulty (PQ2).
For PQ3, we find that 33 participants did not wish for a different ordering of instances (but were still allowed to provide suggestions), four would have preferred an "easy-first", one a "difficult-first", and two an entirely different ordering strategy.
From the 14 free-text answers and feedback via other channels, we identify three general suggestions that may be interesting for future research:
\begin{itemize}
	\item[\textbf{S1:}] Grouping by word rarity.
	\item[\textbf{S2:}] Grouping instances by token overlap.
	\item[\textbf{S3:}] Grouping instances by topic (tweet or alternatively, misconception) similarity.
\end{itemize}
Further analyzing the free-text answers together with the pre-defined answers ("no", "easy-first", "difficult-first", and "other") shows that the participants disagree upon the preferred ordering strategy.
For instance, the participants that suggested S3, disagreed if instances should be grouped by topic similarity to reduce the number of context switches or be as diverse as possible to provide some variety during annotation.
Another five participants (two from Random and one from the other groups each) even explicitly supported a random ordering in the free-text answer.
The disagreement upon the ordering strategy shows the importance of interactively trained estimators that are capable of providing personalized annotation curricula.

\section{Limitations and Future Work}
\label{sec:limitations}
We evaluated AC with an easy-instances-first strategy in simulations as well as in a highly controlled setup using a finite, pre-annotated data set and task-agnostic estimators to minimize possible noise factors. 
To demonstrate the viability of AC with a sufficient number of voluntary annotators, we further chose a dataset that covers a widely discussed topic and manually controlled the annotation difficulty to make it accessible for non-experts.
To evaluate AC with more generalizable results in a real-world scenario, we discuss existing limitations that should be considered beforehand which can also serve as promising research directions for future work.

\paragraph{Difficulty estimators}
Due to novelty of the proposed approach and the lack of well-established baselines, we focused on task-agnostic annotation difficulty estimators such as reading difficulty and annotation time which can easily be applied to a wide range of tasks.
Although our study results show that they work to some extent, our evaluation with existing datasets also shows that especially non-adaptive estimators which approximate the absolute task-difficulty are sensitive to the data domain and annotation task (cf. the low performance of length-based estimators on the SigIE data in \cref{intrinsic}).
Such issues could be addressed by implementing estimators that are more \textit{task-specific}.
For named entity annotations, a general improvement may be achieved by considering the number of nouns within a sentence which can be obtained from a pre-trained part-of-speech tagger.
One may even consider domain-specific word frequency lists to provide a difficulty estimate for entities.
For instance, among the annotated named entities in Muc7\textsubscript{T} "U.S." (occurs 72 times) may be easier to annotate than "Morningstar" (occurs only once); simply based on a word frequency analysis.
Other, more sophisticated approaches from educational research such as item response theory~\citep{baker2001the} and scaffolding~\citep{jackson2020teaching} may also lead to better task-agnostic estimators.
Such approaches and combinations of task-agnostic with task-specific estimators remain to be investigated in future work.

\paragraph{Annotation strategies}
In this work, we focused on developing and evaluating a strategy for our non-expert annotation scenario.
Whereas it proved to be effective in our user study, we also find that our annotators disagree in their preferences with respect to the ordering of instances --- which indicates that investigating \textit{annotator-specific} strategies could be a promising line for future work.
Another shortcoming of the evaluated strategy is that it does not consider an annotator's boredom or frustration~\citep{Vygotsky78}.
Especially when considering larger annotation studies, motivation may become an increasingly important factor with non-expert annotators as they further progress in a task and become more proficient.
Such a strategy may also be better suited for annotation scenarios that involve domain experts to retain a high motivation by avoiding boredom --- for instance, by presenting them with subsequent instances of varying difficulty or different topics.
Domain experts who do not require a task-specific training may also benefit from strategies that focus on familiarizing them with the data domain early on to provide them with a good idea of what kind of instances they can expect throughout their annotations.
To implement strategies that consider annotator-specific factors such as motivation and perceived difficulty, adaptive estimators may have an advantage over non-adaptive ones as they can incorporate an annotator's preference on the fly. 
We will investigate more sophisticated adaptive estimators (also coupled with non-adaptive ones) and strategies in future work and also plan to evaluate AC with domain expert annotators.

\paragraph{Larger datasets}
While using a finite set of annotated instances was necessary in our user study to ensure a proper comparability, AC is not limited to annotation scenarios with finite sets.
However, deploying AC in scenarios that involve a large number of unlabeled instances requires additional consideration besides an annotator's motivation.
In scenarios that only annotate a subset of the unlabeled data (similar to pool-based active learning), an easy-instances-first strategy may lead to a dataset that is imbalanced towards instances that are easy to annotate.
This can hurt data diversity and consequently, result in models that do not generalize well to more difficult instances. To create more diverse datasets, one may consider introducing a stopping criterion (e.g., a fixed threshold) for the annotator training phase and move on to a different sampling strategy from active learning.
Other, more sophisticated approaches would be to utilize adaptive estimators with a pacing function~\citep{kumar-2010} or sampling objectives that jointly consider annotator training and data diversity~\citep{lee-etal-2020-empowering}.
Such approaches are capable of monitoring the study progress and can react accordingly which may result in more diverse datasets.
However, they also face additional limitations in terms of the computational overhead that may require researchers to consider an asynchronous model training in their setup.

\paragraph{Implementation overhead}
Finally, to apply AC in real-world annotation studies, one needs to consider the additional effort for study conductors to implement it. 
Whereas the task-agnostic estimators we provide can be integrated with minimal effort, developing task- and annotator-specific estimators may not be a trivial task and requires a profound knowledge about the task, data, and annotators.
Another open question is how well the time saving of approximately 8--13 minutes per annotator in our study translates to large-scale annotation studies.
If so, then AC could also be helpful in annotation studies with domain experts by resulting in more annotated instances within a fixed amount of time --- however, if not, this would simply lead to a trade-off between the time investment of the study conductor and annotators.
Overall, we find that developing and evaluating further strategies and estimators to provide study conductors with a wide range of choices to consider for their annotation study will be an interesting task for the research community.

\section{Conclusion}
\label{conclusion}
With annotation curricula (AC), we have introduced a novel approach for implicitly training  annotators.
We provided a formalization for an easy-instances-first strategy that orders instances from easy to difficult by approximating the annotation difficulty with task-agnostic heuristics and annotation time.
In our experiments with three English datasets, we identified well-performing heuristics and interactively trained models and find that the data domain and the annotation task can play an important role when creating an annotation curriculum.
Finally, we evaluate the best performing heuristic and adaptive model in a user study with 40 voluntary participants who classified English tweets about the Covid-19 pandemic and show that leveraging AC can lead to a significant reduction in annotation time while preserving annotation quality.

With respect to our initial research questions (cf.~\cref{introduction}), our results show that the order in which instances are annotated can have a statistically significant impact in terms of annotation time (RQ1) and that recent language models can provide a strong baseline to pre-compute a well-performing ordering (RQ2).
We further find that our interactively trained regression models lack adaptivity (RQ3), as they perform well on existing datasets with hundreds or more training instances, but fall behind non-adaptive estimators in the user study.

We conclude that annotation curricula provide a promising way for more efficient data acquisition in various annotation scenarios --- but that they also need further investigation with respect to task-specific estimators for annotation difficulty, annotator-specific preferences, and applicability on larger datasets.
Our analysis of existing works shows that unfortunately, the annotation ordering as well as annotation times are seldomly reported.
In face of the increasing use of AI models in high-stake domains~\citep{sambasivan-2021} and the potentially harmful impact of biased data~\citep{papakyriakopoulos-2020} we ask dataset creators to consider including individual annotation times and orderings along with a datasheet~\citep{gebru2018datasheets} when publishing their dataset.
To facilitate future research, we share all code and data and provide a ready-to-use and extensible implementation of AC in the INCEpTION annotation platform.\footnote{\url{https://inception-project.github.io/}}

\clearpage

\begin{acknowledgments}
This work has been supported by the European Regional Development Fund (ERDF) and the Hessian State Chancellery – Hessian Minister of Digital Strategy and Development under the promotional reference 20005482 (TexPrax) and the German Research Foundation under grant № EC 503/1-1 and GU 798/21-1 (INCEpTION).
We thank Michael Bugert,  Richard Eckart de Castilho, Max Glockner, Ulf Hamster, Yevgeniy Puzikov, Kevin Stowe,  and the anonymous reviewers for their thoughtful comments and feedback, as well as all anonymous participants in our user study.

\end{acknowledgments}

\section*{Ethics Statement}
\label{sec:appendix_ethics}

\paragraph{Informed consent}
Participants of our user study participated voluntarily and anonymously with a self-chosen, unique study key that allows them to request the deletion of their provided data at a later point.
Upon registration, they are informed about the data presented and collected in the study, its further use, and the purpose of the study.
Before collecting any data, participants are explicitly asked for their informed consent.
We do not collect any personal data in our study. 
If participants do not provide their informed consent, their study key is deleted immediately. 
For publication, the study key is further replaced with a randomly generated user id. 

\paragraph{Use of Twitter data}
The CovidLies corpus~\citep{hossain-etal-2020-covidlies} we used to generate the instances for our annotation study consists of annotated tweets. 
To protect the anonymity of the user who created the tweet, we only display the text (removing any links) without any metadata like Twitter user id, or timestamps to our study participants.
We only publish the tweet ids in our study data to conform with Twitter's terms of service and hence, all users retain their right to delete their data at any point.

\clearpage
\starttwocolumn
\bibliography{bibliography}

\end{document}